**Title:**

'Virtual pivot point' in human walking: always experimentally observed but simulations suggest it may not be necessary for stability

**Authors:**

Lucas Schreff [1,2*], Daniel F.B. Haeufle [3,4], Alexander Badri-Spröwitz [5,6], Johanna Vielemeyer [1,7], Roy Müller [1,2]

[1] Department of Neurology/Department of Orthopedic Surgery, Klinikum Bayreuth GmbH, Bayreuth, Germany

[2] Bayreuth Center of Sport Science, University of Bayreuth, Bayreuth, Germany

[3] Hertie Institute for Clinical Brain Research and Center for Integrative Neuroscience, Tübingen, Germany

[4] Institute for Modelling and Simulation of Biomechanical Systems, University of Stuttgart, Germany

[5] Dynamic Locomotion Group, Max Planck Institute for Intelligent Systems, Stuttgart, Germany

[6] Department of Mechanical Engineering, KU Leuven, Belgium

[7] Institute of Sport Sciences, Friedrich Schiller University Jena, Jena, Germany

*Correspondence to: <u>lucas.schreff@gmx.de</u> (Klinikum Bayreuth GmbH; Department of Neurology/Department of Orthopedic Surgery; Hohe Warte 8; 95445 Bayreuth, Germany)

**Abstract**

The intersection of ground reaction forces near a point above the center of mass has been observed in computer simulation models and human walking experiments. Observed so ubiquitously, the intersection point (IP) is commonly assumed to provide postural stability for bipedal walking. In this study, we challenge this assumption by questioning if walking without an IP is possible. Deriving gaits with a neuromuscular reflex model through multi-stage optimization, we found stable walking patterns that show no signs of the IP-typical intersection of ground reaction forces. The non-IP gaits found are stable and successfully rejected step-down perturbations, which indicates that an IP is not necessary for locomotion robustness or postural stability. A collision-based analysis shows that non-IP gaits feature center of mass (CoM) dynamics with vectors of the CoM velocity and ground reaction force increasingly opposing each other, indicating an increased mechanical cost of transport. Although our computer simulation results have yet to be confirmed through experimental studies, they already indicate that the role of the IP in postural stability should be further investigated. Moreover, our observations on the CoM dynamics and gait efficiency suggest that the IP may have an alternative or additional function that should be considered.

# 1 Introduction

Human bipedal locomotion crucially relies on the ability to continuously balance the whole body in an upright configuration. Given the importance of postural stability during walking, various balancing strategies have been investigated with the help of computational models (e.g., Geyer and Herr, 2010; Rummel and Seyfarth, 2010; Sharbafi and Seyfarth, 2015). One such strategy is based on the observed intersection of ground reaction forces (GRF) near a point above the center of mass (CoM) during a single stride (Gruben and Boehm, 2012; Maus et al., 2010). This intersection point (IP) has been interpreted as the pivot point of a virtual pendulum and referred to as the virtual pivot point (Maus et al., 2010). However, human walking cannot accurately be modeled as a pendulum with distributed mass because, e.g., the angular acceleration of a pendulum with distributed mass is 180 degrees out of phase with that of a human body (Gruben and Boehm, 2012). For a rocking rigid body, which does have the phase behavior and an IP above the CoM similar to a walking human, the mass returns to upright without external control, exhibiting stability (Gruben and Boehm, 2012; Lipscombe and Pellegrino, 1993). Humans differ from a single rigid body due to their joints, which they use to move the masses of individual segments. Hence, whether the IP is necessary for postural stability in human walking has yet to be conclusively determined.

So far, in all experimental IP-related human locomotion studies, the GRF intersection point was observed. Examples include: during walking at varying speeds (Gruben and Boehm, 2012; Maus et al., 2010; Vielemeyer et al., 2021), in hip-flexed walking (Müller et al., 2017), while walking and running down visible and camouflaged ground perturbations (Drama et al., 2020; Vielemeyer et al., 2019), and in the gait of patients with Down Syndrome (Vielemeyer et al., 2023). GRFs that do not intersect the CoM have been shown not only in humans but also in quadrupeds (Jayes and Alexander, 1978; Maus et al., 2010), birds (Maus et al., 2010) and bird models (Drama and Badri-Spröwitz, 2020), suggesting common underlying mechanics and control strategies.

To investigate the function of the IP, Maus et al. (2010) developed a computational model with two massless spring-like legs and one trunk segment. They introduced hip moments to ensure that the GRFs intersected at a single point. In other spring-loaded models with a trunk, where the IP was not used as a target variable, it has been shown that the force vectors are focused above the CoM (Rummel and Seyfarth, 2010; Sharbafi and Seyfarth, 2015). Barazesh and Sharbafi (2020) then demonstrated that the physiologically more detailed neuromuscular reflex model of Geyer and Herr (2010) with its default control parameters predicts an IP.

The results and interpretations of previous experimental studies and simulations could indicate that an IP is a prerequisite for stable, steady, and upright walking. In this study, we challenge this indication. We ask if walking without IP is possible and, if so, what effects walking without IP has on robustness and mechanical locomotion efficiency. For our investigation, we applied the neuromuscular walking model of Geyer and Herr (2010) and performed two optimization runs to create gaits with and without IP based on different cost functions. Subsequently, we evaluated the robustness of the different gaits in step-down perturbations and calculated collision fraction values (Lee et al., 2011; Lee et al., 2013) to assess CoM dynamics.

**2 Methods**

We simulated walking with a 2D (sagittal plane) human-like multi-body model (Fig. 1) (Geyer and Herr, 2010) controlled by the neuromuscular reflex control method (Geyer et al., 2003).

We calculated the IP for the single support phase of one stride in steady locomotion. Here, a COM-centered coordinate frame, where the vertical axis is gravity aligned, was used to make it comparable to other studies (e.g., Müller et al., 2017; Vielemeyer et al., 2019). For the calculation of the IP, the GRFs were placed in this coordinate system. Because the GRFs do not intersect at exactly one point, the coefficient of determination $R^2$, derived from a study by Herr and Popovic (2008), helps to assess the spread. $R^2$ reaches a maximum value of 1 and has no lower bound. An $R^2$ value lower than 1 or even negative values indicate that the force vectors are less focused, approaching negative infinity when all GRF vectors are parallel. To determine the IP and the coefficient of determination $R^2$, we adopted the calculation script of Vielemeyer et al. (2021).

To create gaits with and without IP, we applied the covariance matrix adaptation evolution strategy (Hansen, 2006), where we optimized 12 model control parameters. We performed two optimization runs with three-stage cost functions (*J*). As suggested by Song and Geyer (2015) the first two stages were implemented to find stable and steady walking solutions. In contrast to Song and Geyer (2015), we calculated the margin of stability (Hof et al., 2005) for 6 consecutive heel strikes at stage 2. A gait is considered stable and steady when the maximum difference of the 6 values is less than 0.75 cm. To determine the margin of stability, we adapted Schreff et al.`s (2022) calculation script. Only stage 3 differed in the two optimization runs. In the first optimization run, we minimized the coefficient of determination $R^2$ to find a gait without IP (equation 1). In the second run, we maximized $R^2$ to find a gait with IP (equation 2). In addition, we optimized in this run the walking speed ($v_{sim}$). As target speed ($v_{tgt}$) we used the walking speed of the gaits with very low $R^2$ that resulted at the end of the first optimization run (1.25 m/s).

$$J = R^2 \tag{1}$$

$$J = 1 - R^2 + |v_{sim} - v_{tgt}| \tag{2}$$

For both optimization runs, the default control parameters of the model were used as initial parameters. During the optimizations, we saved intermediate results to collect viable gaits with varying $R^2$. Based on a rating of Herr and Popovic (2008) we defined walking with IP when $R^2>0.6$. Lower values characterized walking without IP.

We evaluated the robustness of all gaits for which we stored the underlying control parameters during the two optimization runs by testing their ability to recover from step-down perturbations. After each successful trial, the step-down height *h* was increased by 1 cm until the model could no longer recover from the perturbation with its respective control parameters. For more information see Haeufle et al. (2018) and Schreff et al. (2022).

To evaluate the efficiency of CoM dynamics during walking, we performed a collision analysis (more detailed explanation in Fig. 1) based on collision angle (CA) and fraction (CF) equations established by (Lee et al., 2011; Lee et al., 2013). Importantly, Lee et al. (2011) show that high collision angle and, consequently, collision fraction values correspond to a high mechanical cost of transport. Here, to assess the efficiency of our optimized gaits, we investigate whether there is a correlation between the CF and $R^2$.

The equations for the collision analysis, as well as for the calculation of the coefficient of determination and the margin of stability can be found in the supplementary information.

We optimized the model's control parameters running Matlab® Simulink® R2021a, with the ode15s solver, a maximum step size of 10 ms, and relative and absolute error tolerances of $10^{-3}$ and $10^{-4}$, respectively.

## 3 Results

We optimized gaits for high and low $R^2$ in two independent runs. With both cost functions, the model produced stable walking patterns, with $R^2$ values ranging from -835.08 to 0.93. We named the gait with the highest $R^2$ value the "IP gait", with a walking speed of 1.25 m/s and a step length 0.78 m. For the lowest $R^2$ "non-IP gait", the model walked at a speed of 1.24 m/s with a step length of 0.80 m. In comparison, the "default gait" with identical parameters as the model of Geyer and Herr (2010) has an $R^2$ of 0.83, at a walking speed of 1.36 m/s and a step length of 0.77 m. Fig. 2 shows the IP plots of the three different gaits. In the supplementary videos 1, 2 and 3, the animations of the gaits can be seen.

Fig. 3 illustrates parameters relevant to establish an IP. The three gaits differ mainly in the CoM trajectories in the vertical direction and their GRFs. For the non-IP gait, more oscillations are visible for the vertical GRFs, compared to the default gait. The horizontal GRFs of the non-IP gait stand out as they are mostly positive during the single support phase (non-semi-transparent blue line of Fig. 3d).

We examined all gaits of the two optimization runs for robustness against step-down perturbations and collision potential. The results can be seen in Fig. 4. For robustness, there is no clear trend in the gaits with IP ($R^2>0.6$) and without IP ($R^2<0.6$). However, it seems that CF values increase for gaits without IP.

Due to the choice of our cost functions, it cannot be excluded that non-IP gaits with low CF values exist. At this point, we performed a further optimization run in which we minimized CF values in addition to Eq. 1. However, we could not find any gaits with $R^2<-1$ that featured CF values below 0.6.

## 4 Discussion

The optimized gaits of the neuromuscular reflex model predict that stable, steady, and upright walking is possible without the IP-typical intersection of GRFs. While our simulation results on the relation between IP and stability have yet to be confirmed in human experimental investigations, e.g., by "exaggerated walking" as investigated by Herr and Popovic (2008), we show that an IP is not necessary for simulation of stable bipedal locomotion.

In previous studies, it was assumed that the IP is necessary for postural stability. Maus et al. (2010) evaluated the influence of the IP on postural stability using a step-down perturbation with a very low height of 5 mm. In contrast, we used perturbation heights of at least 1 cm for our investigations. The model can recover from ground drops of up to 3 cm, with default control parameters. The IP and non-IP gaits reject step-downs mainly in the range between 1 cm and 5 cm. In sum, we found no correlation between $R^2$ and the manageable obstacle height (Fig. 4a). We conclude that an IP is not necessary for robustness against step-down perturbations, respectively postural stability during walking.

In our simulations, we used the default gait as a reference because it accurately predicts the kinematic and kinetic data (for more information see Geyer and Herr, 2010) known from experimental studies on human gait (e.g., Aminiaghdam et al., 2017; Vielemeyer et al., 2021). In comparison, the horizontal and vertical GRFs of the non-IP gait exhibit more oscillations. Furthermore, high collision fraction values (up to 0.82, Fig. 4b) indicate that the CoM dynamics of non-IP walking gaits have increased potential for collisions, leading to a higher mechanical cost of transport and lower locomotion efficiency. However, some IP gaits with very high $R^2$ values (0.91, Fig. 4b) also show increased CF values (0.66), suggesting that IP walking is not a guarantee for avoiding collisions and a minimal cost of transport. Through our additional investigations, we found that walking with low collision potentials, which means efficient walking, seems possible only for gaits with high $R^2$.

In summary, we predict that stable and upright human walking without an IP should be possible. However, we cannot rule out that the IP plays a role in promoting stability during locomotion, because it is possible that the absence of an GRF intersection point in the non-IP gaits and therefore a loss of the IP's potential stabilizing effect can be compensated by another stabilizing strategy. Given the many experimental studies demonstrating focused GRF in human walking, we challenge the community to further investigate the function of the IP. Our findings on the CoM dynamics and gait efficiency suggest that the IP may have an alternative or additional function that should be taken into consideration.

**Data availability statement**

The datasets generated and analyzed during the current study are available from the corresponding author on reasonable request.

**Conflict of interest statement**

The authors declare that no financial and personal relationships with other people or organizations have inappropriately influenced the content of the work reported in this paper.

**Acknowledgements**

This work was funded by the Deutsche Forschungsgemeinschaft (DFG, German Research Foundation) 327485414, 449912641.

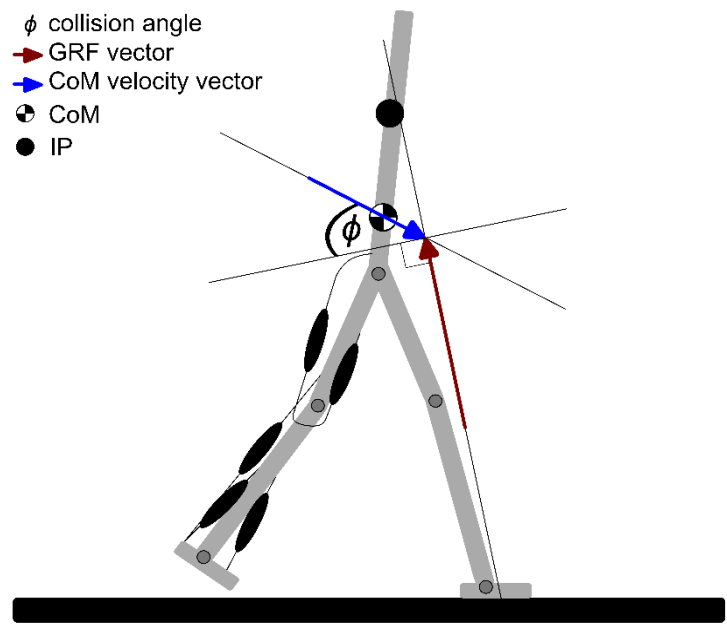

Fig. 1. The used neuromuscular reflex model (Geyer and Herr, 2010) considers seven segments, all with anthropomorphic masses (1 upper body with 53.5 kg, 2 thighs 8.5 kg each, 2 shanks 3.5 kg each, and 2 feet 1.25 kg each), connected by hinge joints. Each leg includes seven Hill-type muscle-tendon units (the gluteus and hip flexor muscles are not shown). Stimulation patterns of the muscles are generated by reflex-based signals, mainly from proprioceptive muscle force and length feedback. To balance the trunk, muscle stimulations of the hamstrings, hip flexors, and gluteus additionally depend on the trunk's forward lean angle and velocity. The GRF line-of-action passes close to the IP located above the CoM of the model. To illustrate the collision-based analysis (Lee et al., 2011; Lee et al., 2013), the collision angle is the angle between the CoM velocity vector and the perpendicular to the GRF vector (eqns. 1.6, 1.7, Lee et al., 2013). The collision fraction (min 0, max 1) is then the weighted average of the instantaneous ratio of actual collision (collision angle) to the potential collision (eqn. 1.14, Lee et al., 2013). Lee defines the potential collision as the sum of the instantaneous angles of the GRF (with respect to vertical, $\theta$) and the velocity vector (with respect to horizontal, $\lambda$) (page 4 of Lee et al., 2013). A zero collision fraction is found in a rolling wheel where the wheel's CoM velocity vector and its GRF vector are oriented perpendicularly (Lee et al., 2011). A high collision fraction indicates that CoM velocity and GRF vectors feature, in average, a more non-perpendicular orientation during the stance phase.

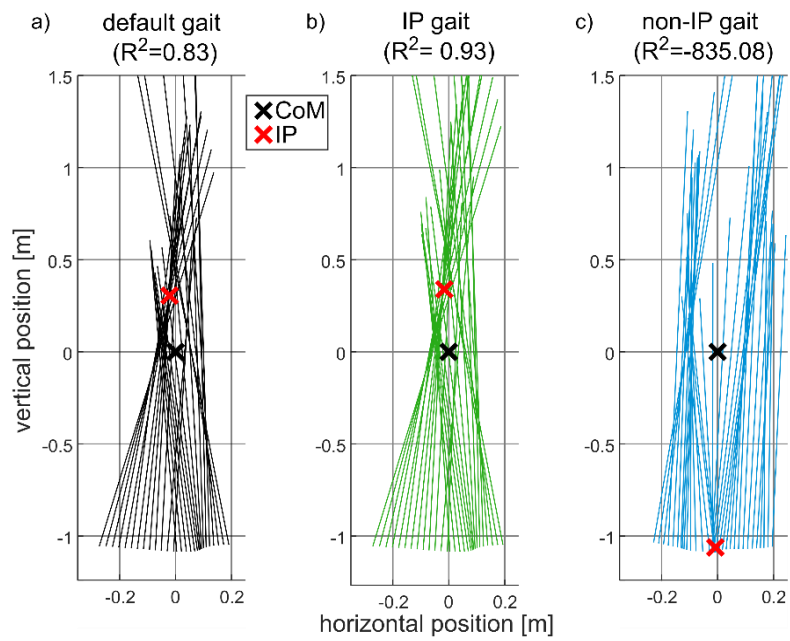

Fig. 2. Ground reaction force plots in a CoM-centered coordinate system for three gaits: default (Geyer and Herr, 2010) (a), IP (b), and non-IP (c).

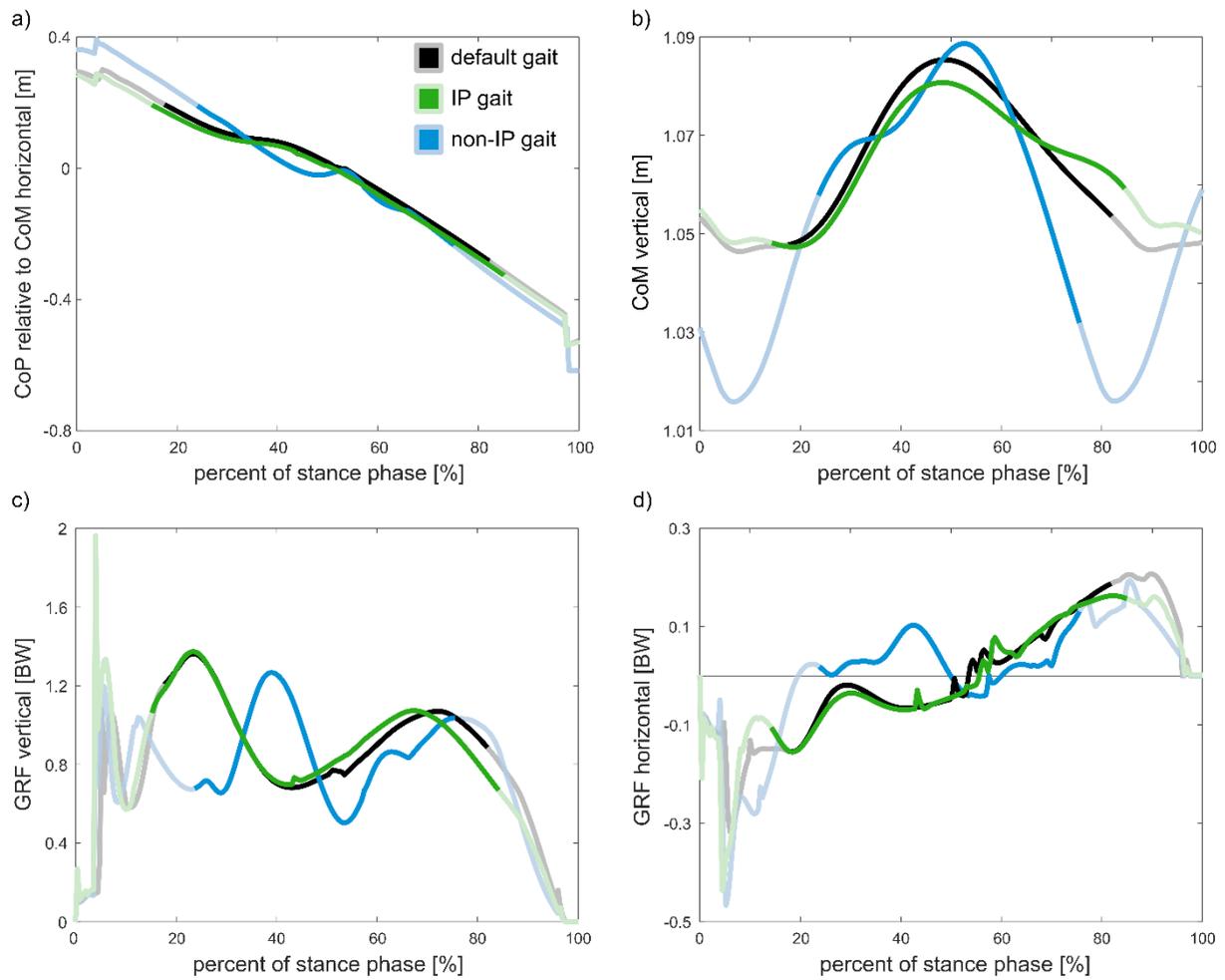

Fig. 3. The center of pressure (CoP) relative to the horizontal CoM position (a), vertical CoM position (b), vertical GRFs (c), and horizontal GRFs (d) plotted over an entire stance phase. Shown are the default, the IP, and the non-IP gait. The IP calculation is conducted during the single-support phase plotted here with non-transparent lines. Semi-transparent lines indicate the double-support phase.

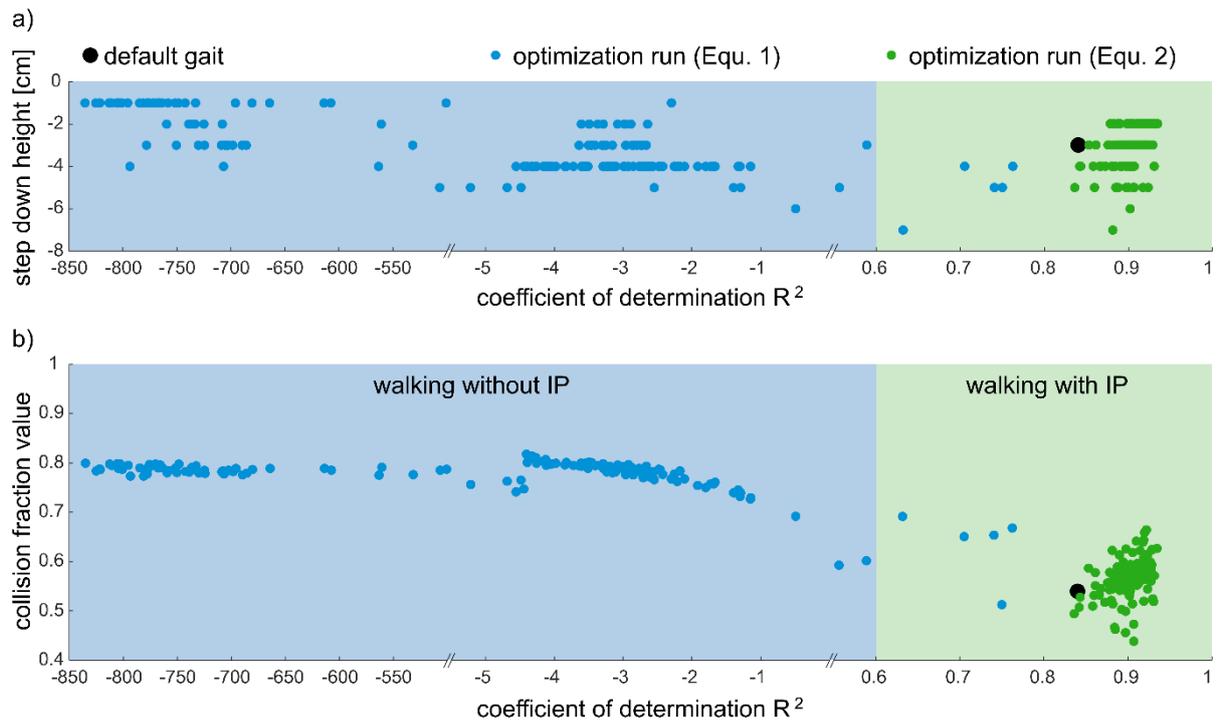

Fig. 4. Maximum step-down heights (a) and collision fraction values (b) of all optimized and stable gaits. Green background areas indicate IP walking gaits with $R^2$ values above 0.6. Blue areas indicate non-IP walking gaits, with $R^2$ values below 0.6. The default gait (Geyer and Herr, 2010) is marked as a black data point.